\documentclass[submission,copyright,creativecommons]{eptcs}
\providecommand{\event}{AREA 2020} 
\usepackage{breakurl}             
\usepackage{underscore}  
\usepackage[table]{xcolor}
\usepackage{subcaption}
\usepackage[labelformat=parens,labelsep=quad,skip=3pt]{caption}
\usepackage{cite,xcolor,textcomp,graphicx,tabularx, algorithmic,amsmath,amssymb,amsfonts,hyperref,todonotes,dsfont,tabularx,booktabs,hyperref,cleveref,tikz,marvosym,multirow,textcomp}
\title{Co-Simulation of Human-Robot Collaboration: from Temporal Logic to 3D Simulation}
\author{Mehrnoosh Askarpour
\institute{McMaster University\\ Canada}
\email{askarpom@mcmaster.ca}
\and
Matteo Rossi \qquad\qquad Omer Tiryakiler
\institute{Politecnico di Milano\\
Italy}
\email{\quad matteo.rossi@polimi.it \quad\qquad omer.tiryakiler@mail.polimi.it}
}
\newcommand\resq[1]{
\noindent 
\\
\fcolorbox{black!40!black}{black!5}{
\noindent 
 \parbox{0.98\columnwidth}{\noindent  #1}}\\
}
\def\titlerunning{Co-Sim. of HRC: From Temporal Logic to 3D Simulation}
\def\authorrunning{M. Askarpour, M. Rossi \& O. Tiryakiler }
\newcommand{\ExternalLink}{%
    \tikz[x=1.2ex, y=1.2ex, baseline=-0.05ex]{%
        \begin{scope}[x=1ex, y=1ex]
            \clip (-0.1,-0.1) 
                --++ (-0, 1.2) 
                --++ (0.6, 0) 
                --++ (0, -0.6) 
                --++ (0.6, 0) 
                --++ (0, -1);
            \path[draw, 
                line width = 0.5, 
                rounded corners=0.5] 
                (0,0) rectangle (1,1);
        \end{scope}
        \path[draw, line width = 0.5] (0.5, 0.5) 
            -- (1, 1);
        \path[draw, line width = 0.5] (0.6, 1) 
            -- (1, 1) -- (1, 0.6);
        }
    }
\newcommand{\irr}[1]{\cellcolor{gray!35}{#1}}
\newcommand{\unsat}{$\mathsf{UNSAT}$}
\begin{document}
\maketitle

\begin{abstract}
Human-Robot Collaboration (HRC) is rapidly replacing the traditional application of robotics in the manufacturing industry. Robots and human operators no longer have to perform their tasks in segregated areas and are capable of working in close vicinity and performing hybrid tasks---performed partially by humans and by robots.

We have presented a methodology in an earlier work~\cite{tro} to promote and facilitate formally modeling HRC systems, which are notoriously safety-critical. Relying on temporal logic modeling capabilities and automated model checking tools, we built a framework to formally model HRC systems and verify the physical safety of human operator against ISO 10218-2~\cite{ISO:10218-2} standard. In order to make our proposed formal verification framework more appealing to safety engineers, whom are usually not very fond of formal modeling and verification techniques, we decided to couple our model checking approach with a 3D simulator that demonstrates the potential hazardous situations to the safety engineers in a more transparent way. This paper reports our co-simulation approach, using Morse simulator~\cite{morse} and Zot model checker~\cite{zot}.
\end{abstract}

\section{Introduction}
Not long ago, robotic systems in manufacturing industry were restricted to work apart from human operators, and thus benefiting from both human flexibility and machine efficiency was indeed a challenge. Human operators were not allowed to enter in the workspace of the robot, which was usually fenced, unless the robot was fully motionless.

Nowadays, though, things are quickly moving towards close interactions between human and robot and their easy and smooth collaboration. The resulting field of these changes is referred to as Human-Robot Collaboration (HRC), which despite apparent benefits, introduces new safety concerns due to close vicinity and frequent contact with human operators. 

Our previous work~\cite{tro} proposes a formal verification based methodology for safety assessment of industrial HRC systems. We generated a framework that contains formally modeled modules to replicate (i) main elements of HRC systems (i.e., the human, the robot, the layout of the workspace, and the executing task), (ii) physical hazard definitions from ISO 10218-2~\cite{ISO:10218-2}, and (iii) risk estimation procedure from ISO 14121~\cite{ISO:14121}. Whilst (ii) and (iii) are fully reusable, (i) needs to be customized for each new scenario based on its specification. However, the framework provides an easily customizable and modifiable template for the safety assessor. The model is created by TRIO temporal logic \cite{furia2012modeling} in order to better capture the temporal dynamics of the system which could consequently affect safety. For example, if human and robot both arrive at point $x$ at the same time, a hazardous contact could happen; while if one reaches point $x$ and remains firm until the other one arrives that contact could be negligible. TRIO formulae use temporal modal operators to describe quantitative temporal requirements which are thoroughly explained in \cite{furia2012modeling}. For example, given that $\phi$ is a proposition and d is a constant value, $\mathsf{Dist}(\phi,d)$ holds at time $t$ if, and only if, $\phi$ holds at time $t+d$ and $\mathsf{Alw}(\phi)$ means that $\phi$ always holds (i.e., $\forall t(\mathsf{Dist}(\phi,t))$).

\begin{figure}
    \centering
    \includegraphics[width=.8\textwidth]{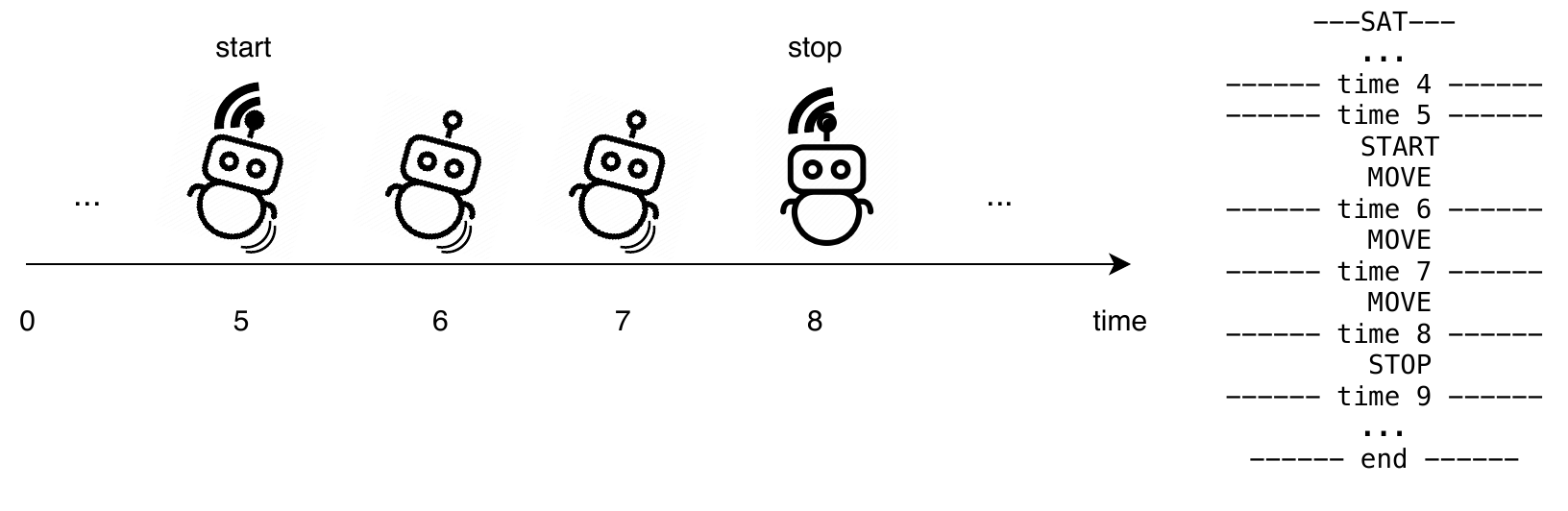}
    \caption{On the right the output of Zot is demonstrated for a simple robot movements formalized in formula\ref{eq:move}. The robot receives signals $start$ at $t = 5$ and $stop$ at $t = 8$, therefore it is moving at $t = 5, 6, 7$ and still at $t = 8$. The figure on the left visualizes the Zot output for easier understanding.}
    \label{fig:hist}
\end{figure}

The model then is checked against a safety property (i.e., the risk value estimated for any identified hazard is not greater than a tolerable threshold) by an automated model checker, called Zot~\cite{zot}. In particular, Zot is a bounded satisfiability checker that as input receives temporal logic models and implements several techniques for checking the satisfiability of those models in an automated manner \cite{PMS13,BPR15}. More precisely, given an input a temporal logic formula $\phi$, Zot returns either a history satisfying $\phi$ or the value \unsat, which stands for ``unsatisfiable''. The output of Zot in the former case is a textual explanation of a trace of a model in a bounded time interval. For example, if Zot evaluates the model for a time bound of 30 time instants, it produces a set of all the values of the models' variables and logical predicates for each time instant (30 sets). Zot shows comparable performance with respect to well-known bounded model checkers such as NuSMV \cite{DBLP:conf/cav/CimattiCGGPRST02}.
\Cref{fig:hist} shows an example of the output of Zot after verifying a simple robot movement model. The robot regularly receives two signals $start$ and $stop$ which are apart by three time instants; $start$ and $stop$ could never happen at the same time. 
This behaviour is modeled by the TRIO formula below
and visualized in \cref{fig:hist}(left). On the \cref{fig:hist}(right), the output of Zot for this example could be found.
\begin{equation}
\mathsf{Alw}(start \Rightarrow \mathsf{Dist}(stop , 3) \wedge \neg (start \wedge stop))
\label{eq:move}
\end{equation}

Without loss of generality, we decided to keep the model discrete so not to loose the decidability of the model which means:
\begin{itemize}
    \item Time is not continuous and is considered as a set of time instants.
    \item The model of the human and the robot consists of discrete points of interests on their body. The attachment and dynamics of these points are reproduced via logic formulae as a reusable template.
    \item The layout is also discretized and considered as a grid of locations that are not necessarily of the same size and shape. Consequently, movement is formalised as changing the current position from one location to an adjacent one. The varying grid sizes are due to the characteristics of each location; we put together close by areas that have similar characteristics in terms of safety. It reduces the complexity of the layout model and, in turn, avoids state-space explosion for large layouts.
\end{itemize}
More details on the discretized could be found in \cite{tro}.
With an incremental approach, each time the safety property is violated the model needs to be improved by the safety assessor and re-verified. The modifications are in form of risk mitigations such as ``robot slows down as hazard x is detected", ``robot retracts as hazard y is detected".

In order for the safety assessor to better reflect on the hazardous situations and define suitable risk mitigations for them, we want to integrate the results of formal verification with a 3D simulator. It helps the safety assessor to detect each hazard and its cause(s), or check the effectiveness of a new risk mitigation. 
This indeed is a co-simulation approach were the safety assessor is provided with both precision of 3D simulation and exhaustiveness of model checking. A similar approach is used by Webster et. al\cite{doi:10.1177/0278364919883338} where the underlying formal model is probabilistic timed automata. However, our underlying formal verification approach instead is based on temporal logic formulae, which are compatible with Zot.

The present paper reports our co-simulation approach with Morse 3D simulator~\cite{morse}, chosen based on a set of factors that will be discussed in the rest of the paper. The paper is structured as follows: \Cref{sec:sim} reviews the state of the art simulators, \Cref{sec:res} shows our experimental results, and finally \Cref{sec:con} concludes.

\section{Choosing a 3D Simulator}
\label{sec:sim}
This section reports four key factors that we took into account for choosing an appropriate simulator to be coupled with our formal methodology in the following.

\paragraph{Community:} We need to integrate a simulator in our framework which is already firmly established, has a vibrant community of users, and has no indication of abandonment or decay. Tools with a large user community are notoriously more reliable and offer more support via Q\&A platforms, hence are easy to use and learn tools.

\paragraph{Integration Capability:} It is also important for a simulator to be cross platform and portable, and support industry-standard protocols and communication methods known for robotic applications (e.g., Robot Operating System ROS~\cite{ros}, YARP~\cite{yarp}, COLLADA~\cite{collada}).

\paragraph{Licensing (Open-source):} As we want our tool to be publicly accessible rather than being connected to a proprietary, corporately-owned simulator, we look for simulators which are free to use (at least non-commercially) without any restrictions on the duration of use or the tools available.

\paragraph{Human Modeling Capability:} Human agent is most certainly an important part of HRC systems and strongly influences safety. Thus, it is essential for the simulator we choose to be capable of simulating human behaviour. Please note here that we are interested in simulating the physical involvement of human and robot while performing their tasks, and by simulating human we only means the observable physical manifestation of human behaviour (e.g., moving, picking, placing).

\begin{table}[t!]
\caption{Important factors of a 3D simulator as a co-simulation tool for HRC formal models.\newline
-: Not verifiable \hspace{4cm} \texttimes: Not having the desired factor
\newline
\checkmark: Having the desired factor \hspace{2cm} Text: Detail on what is supported by the desired factor
}
\begin{tabularx}{\columnwidth}{X|X|X|X|X|X}
\toprule
& Community& \multicolumn{2}{c|}{Integration Capability}& Open-source& Human Modeling Capability
\\
\cline{3-4}
&  & Portable & Standard Protocol &  & 
\\
\hline
Gazebo \href{http://gazebosim.org/}{\ExternalLink}&\checkmark&not for Windows and Mac& ROS/ YARP &\checkmark&\checkmark 
\\
\hline
OpenRAVE \href{http://openrave.org/}{\ExternalLink}& \texttimes & not for Mac& COLLADA/ ROS& \checkmark & \texttimes
\\
\hline
\irr{RoboDk} \href{https://robodk.com/index}{\ExternalLink}&\irr{\texttimes}&\irr{-}& \irr{-} &\irr{Proprietary }&\irr{\texttimes}
\\
\hline
CoppeliaSim (V-REP) \href{http://www.coppeliarobotics.com/}{\ExternalLink}& \checkmark &\checkmark & ROS &\checkmark & \checkmark
\\
\hline
Morse \href{https://www.openrobots.org/morse/doc/stable/morse.html}{\ExternalLink}& \checkmark & not for Windows& ROS/ YARP
& \checkmark & \checkmark
\\
\hline
Webots \href{https://cyberbotics.com/}{\ExternalLink}& \texttimes & \checkmark & ROS & \checkmark & \checkmark
\\
\hline
\irr{Actin} \href{https://www.energid.com/actin}{\ExternalLink}&\irr{\texttimes}&\irr{\checkmark}&\irr{ROS} &\irr{Proprietary}  &\irr{\texttimes}
\\
\hline
\irr{KUKA.Sim} \href{https://kuka.com}{\ExternalLink} & \irr{\texttimes} & \irr{-} & \irr{-} & \irr{Proprietary} &\irr{\texttimes} 
\tabularnewline 
\bottomrule 
\end{tabularx}	
\label{Tab:sims}
\end{table}
\Cref{Tab:sims} summarises our analysis on seven most cited simulators in academic and industrial world. We have categorized them based on the four key factors already discussed and found out that simulators with more of an industrial agenda unfortunately are not open-source. Thus, using them makes it difficult for other researchers to reproduce and study our work. Moreover, in some cases not even a trial version is accessible so it is not even possible to verify their integration capability.
The gray boxes refer to this group of simulators that we basically had to exclude from our further considerations, and the unverifiable factors have been highlighted with a ``$-$" sign.

As the table shows in the first column, Gazebo, CoppeliaSim and Morse have the bigger community and more reliable documentation and have the capability to model human agent. Noori et al. performed a comparison between Gazebo and Morse \cite{8088134} and concluded that Morse performs better for larger test cases with multiple robot agents.
The competition between CoppeliaSim and Morse remains very close. Even though CoppeliaSim has stronger portability, we decided to proceed with Morse due to more specific HRC examples \cite{10.1007} of its usage compared to more general robotic cases with CoppeliaSim. We figured that, unlike several consecutive problems that portability causes for different simulators \cite{afzal2020study}, luckily the portability issue of Morse for windows is resolvable via a Docker image or a virtual machine.
\section{Co-Simulation Implementation and Results}
\label{sec:res}
\begin{figure}[t]
    \centering
    \includegraphics{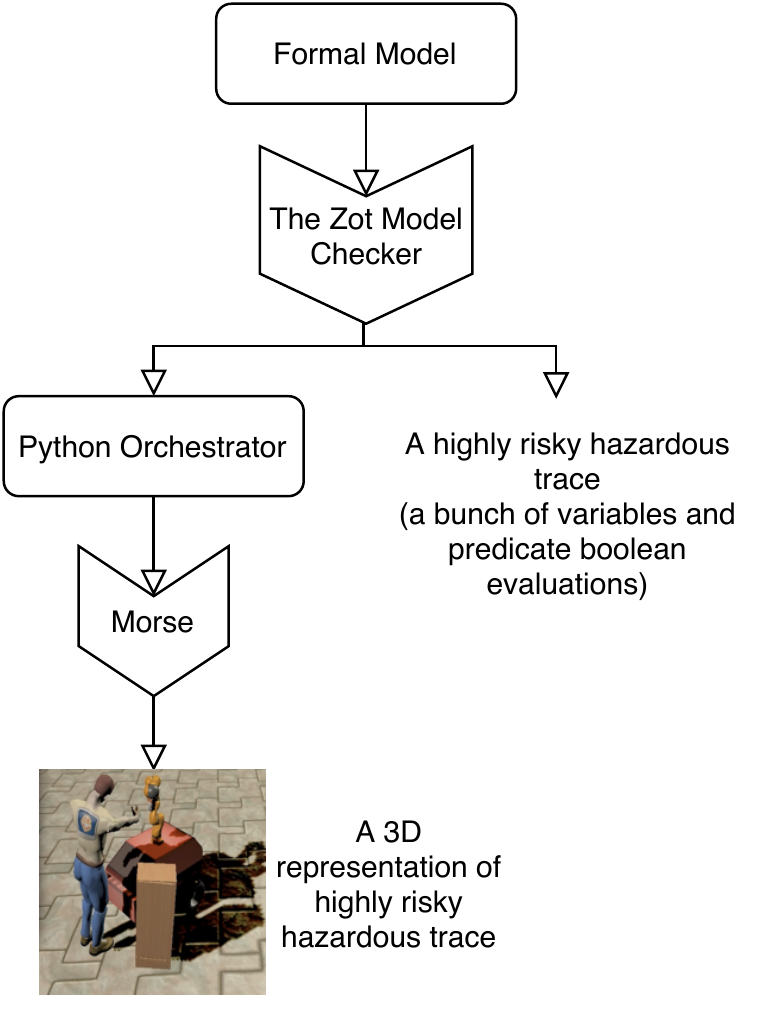}
    \caption{The overview of our co-simulation approach. The rectangles introduce our products, and chevrons represent off-the-shelf tools we used.}
    \label{fig:overview}
\end{figure}

As shown in \cref{fig:overview}, we have implemented an orchestrator using python API of Morse, that works as a middleware between the Zot model checker and the Morse Simulator. It grabs the output information of Zot and translates them to standard commands of Morse. 
The orchestrator then reads this output, recognizes the required commands to be sent to the simulator, generates those commands, and runs them for each time instant. For example, assuming that variable $p_x$ replicates the position of the robot in a discretely modeled layout, the orchestrator reads the value of $p_x$ at every time instant; if $p_x$ at time instant $t$ is different from its value at instant $t + 1$, the orchestrator sends motion command with the two values of $p_x$ at $t$ and $t + 1$ as source and destination of the move.

We generated the formal model for two case studies, of which one is reported here, and ran the formal verification, and then tried our new orchestrator to see if its results are the same as the one reported by the model checker. More precisely, our objective was to make sure if the simulator displays the same hazards as the model checker, and were looking for physical contacts on the 3D images that looked dangerous or hazardous but have not been reported as such by the model checker. We also were looking for cases that are reported by either model checker or simulator and not the other one. Namely, we wanted to see which one catches the most hazardous situations and which one has less false positive reported cases.
We have used an industrial manufacturing task as our test case in which a KUKA manipulator arm, mounted on a moving platform, needs to grab a piece from a given position, move to where a human is standing and then hand them the grabbed piece. The human is free to move around while robot is moving and thus contacts are very much possible.
We ran the verification and simulation multiple times and made the following observation.

\begin{figure}
  \begin{subfigure}{.5\columnwidth}
    \centering\includegraphics{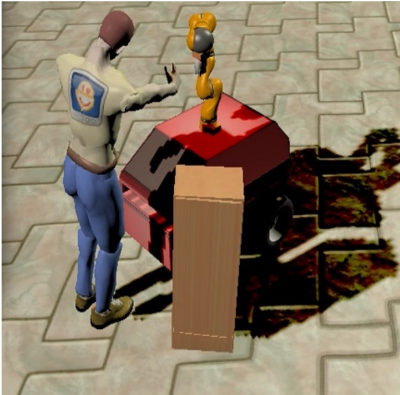}
    \caption{Morse output for a single time instant, where the gripper of the robot and human arm are very close but an actual contact has not happened.}
    \label{fig:exp1}
  \end{subfigure}
  \hspace{.1cm}
  \begin{subfigure}{.5\columnwidth}
    \centering\includegraphics[scale=.42]{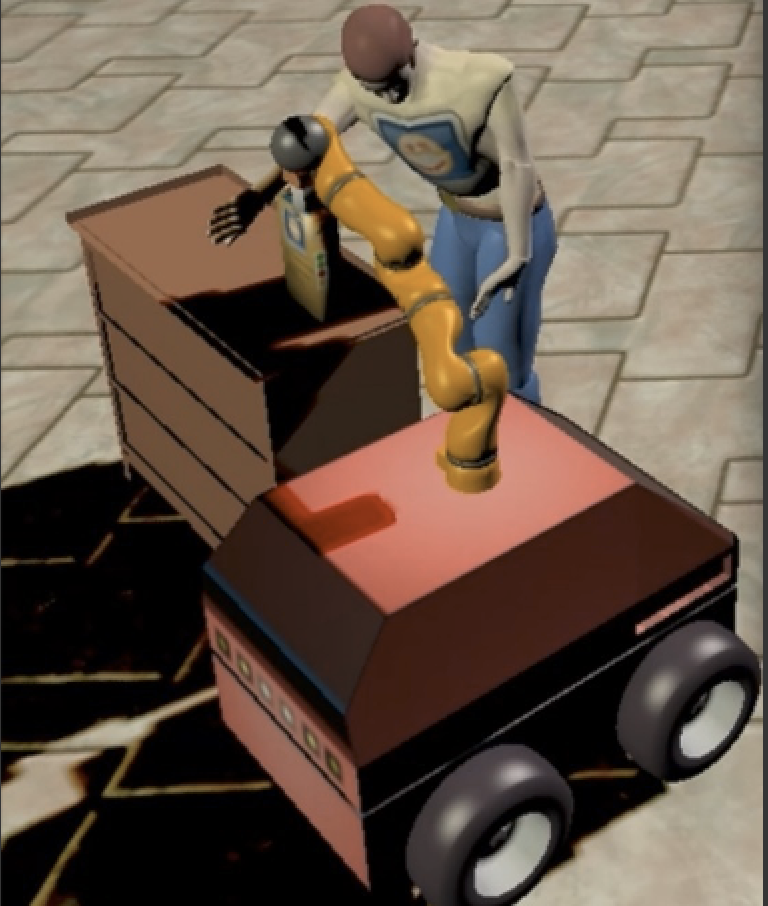}
    \caption{Morse output for another time instant, where model checker and simulator both detect a hazardous contact between human arm and the gripper of the robot.}
    \label{fig:exp2}
  \end{subfigure}
  \caption{Morse output examples.}
\end{figure}

\resq{There are hazardous situations detected by the model checker which are not depicted as such with the simulator, and thus could be perceived as false positives. \Cref{fig:exp1} shows an example of Morse output, where the situation seems to be harmless but the model checker reported a hazard. We suspect that the problem is due to the discrete definition of the layout in the formal model. Basically, when the robot gripper $p_g$ and the human arm $p_a$ are at the same location ($p_g = p_a = l_x$) the model checker recognizes them as being very close and possibly touching each other. However, the simulator shows that it might not necessarily be the case. This could be the additional precision that 3D simulation could add to the model checking results.}

Other than the observation just made, all of the time instants had the same outcome with both of the tools, like the one shown in \cref{fig:exp2}.
\section{Conclusion and Discussions}
\label{sec:con}

This paper is a report of our work on integrating a 3D simulator called Morse within our formal verification framework for HRC systems, and developing a co-simulation approach.
We described our analysis of state of the art robotic simulators and why we chose Morse, its integration within our framework, and obtained results on one of our experimental examples.

Our objective was to use simulation as a complement to model checking in order to combine the precision of the former and exhaustiveness of the latter. In fact, the experiments show that 3D simulator would help us to identify the hazardous situations that are falsely identified by the model checker. 
Implementing risk mitigations in HRC systems is sometimes very expensive (e.g., add emergency stop buttons or wearable signal emitters), or would compromise system efficiency (e.g., robot should slow down until an external event or signal is received). Thus, identifying falsely identified hazards before the physical deployment of a robotic system of interest and setting its workspace would be very profitable for system designers and safety assessors.

\begin{figure}
    \centering
    \includegraphics[scale=.5]{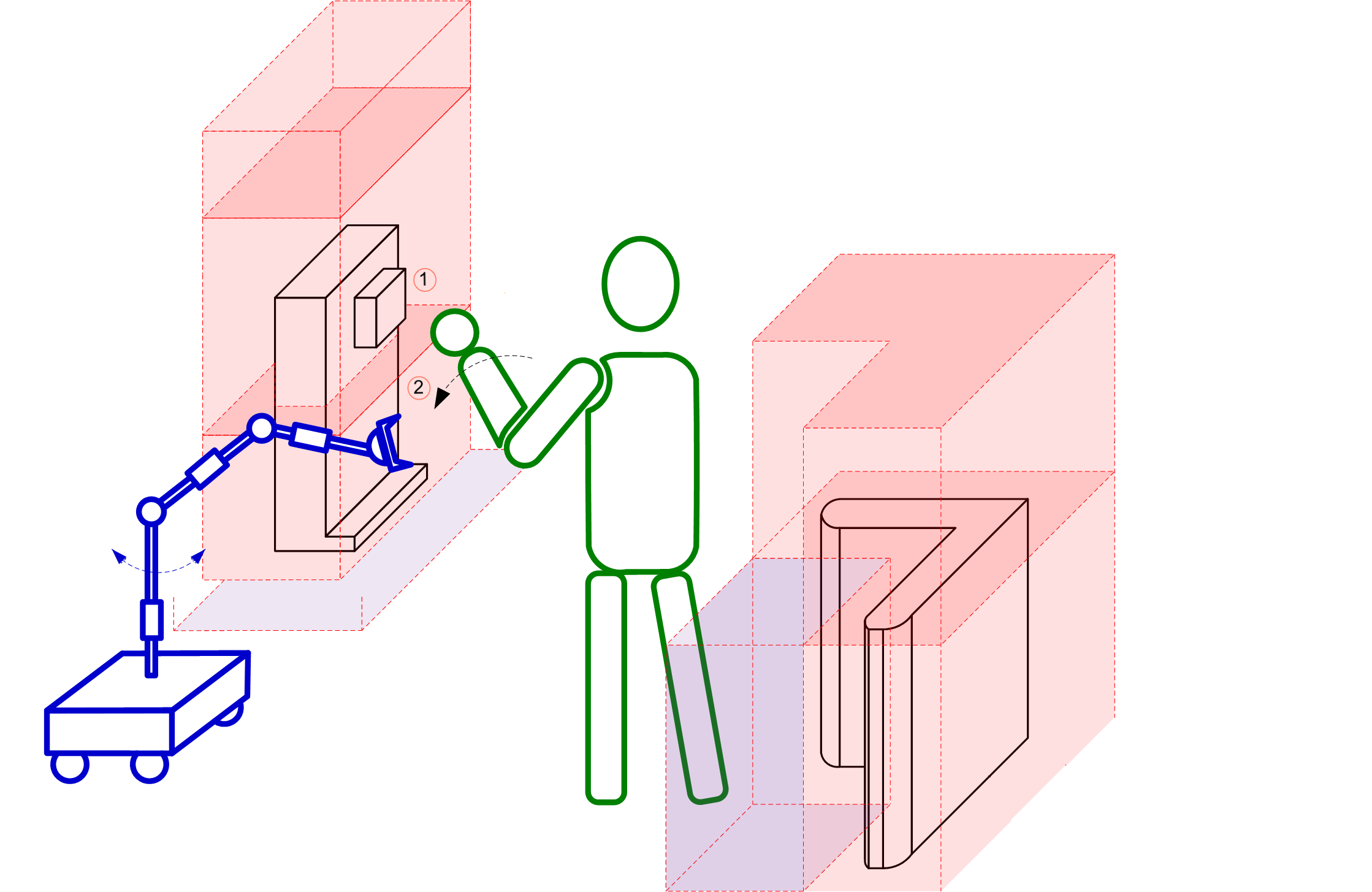}
    \caption{A partial example of discretization of the layout into three dimensional cubic areas.}
    \label{fig:layout}
\end{figure}

There is however a plausible problem with simulating a discrete formal model with a 3D simulator. The discrete definition of the layout makes the translation of a formal model to a 3D view not fluent. Movement in formal model are defined as moving from one location to another, and the bigger these locations are the less smoothly movements will be simulated. For example, \cref{fig:layout} shows a partial discretization of a workcell into 3 dimensional cubic areas which could be mapped into the simulation environment. If the gripper needs to move from the location highlighted as (1) to the location highlighted as (2), the temporal formal model captures this move by assigning a time bound to it (e.g., if the gripper is moving from (1) to (2), then it is moving out of (1) for three time instants and then it will be in (2)). However, the simulator would depict a sudden jump from (1) to (2) after three snapshots. This will create a simulated trace of the system which does not seem very smooth.
The problem becomes negligible when the layout locations (pink cubic areas) are small enough. However, this could cause a huge increase in the size of the model, and consequently the famous state-space explosion with the model checker. Hence, as a future work we are planning to inspect what is the best layout dicretization scale that provides a trade-off between the size of the model and the smoothness of the movements shown by the simulator. 
Moreover, currently the human replica in the formal model is more complex than the simulated one. Formal model has an additional module that, if included in the verification, is able to reproduce hazardous situations caused by human erroneous behavior\cite{ASKARPOUR2019465}. We are planning to dedicate a series of co-simulation experiments to this module of the formal model and identify missing situations from the formal verification process that could be revealed by simulation.
We are also open to the idea of trying out other simulators such as V-rep to see if they could improve the movement issue with different motion controlling commands.

\bibliographystyle{eptcs}
\bibliography{bib}
\end{document}